\documentclass[journal]{IEEEtran}

\usepackage{times}
\usepackage{epsfig}
\usepackage{graphicx}
\usepackage{amsmath}
\usepackage{amssymb}
\usepackage{graphicx,subfigure}
\usepackage[ruled]{algorithm2e}
\usepackage{algorithmic}
\usepackage{multirow}
\usepackage{xcolor}
\usepackage{epstopdf}
\SetKwRepeat{doWhile}{do}{while}
\DeclareMathOperator*{\argmax}{argmax}

\begin{document}

\title{Interest Point Detection based on Adaptive Ternary Coding}

\author{Zhenwei~Miao,~\IEEEmembership{Member,~IEEE},  Kim-Hui Yap,~\IEEEmembership{Senior Member,~IEEE} and Xudong~Jiang,~\IEEEmembership{Senior Member,~IEEE }
\thanks{This research was carried out at the Rapid-Rich Object Search (ROSE) Lab at the Nanyang Technological University, Singapore.  The ROSE Lab is supported by the National Research Foundation, Prime Minister’s Office, Singapore, under its IDM Futures Funding Initiative and administered by the Interactive and Digital Media Programme Office.}
\thanks{The authors are with the School of Electrical and Electronics Engineering and with Rapid-Rich Object Search (ROSE) Lab, Nanyang Technological University, 639798 Singapore. (e-mail:
miaozhenwei@gmail.com; exdjiang@ntu.edu.sg; ekhyap@ntu.edu.sg).}
}


\maketitle

\begin{abstract}
In this paper, an adaptive pixel ternary coding mechanism is proposed and a contrast invariant and noise resistant interest point detector is developed on the basis of this mechanism. Every pixel in a local region is adaptively encoded into one of the three statuses: bright, uncertain and dark. The blob significance of the local region is measured by the spatial distribution of the bright and dark pixels. Interest points are extracted from this blob significance measurement. By labeling the statuses of ternary bright, uncertain, and dark, the proposed detector shows more robustness to image noise and quantization errors. Moreover, the adaptive strategy for the ternary cording, which relies on two thresholds that automatically converge to the median of the local region in measurement, enables this coding to be insensitive to the image local contrast. As a result, the proposed detector is invariant to illumination changes. The state-of-the-art results are achieved on the standard datasets, and also in the face recognition application.
\end{abstract}


\IEEEpeerreviewmaketitle

\section{Introduction}
 \label{sec:intro}

\IEEEPARstart{A}~well-designed interest point detector is supposed to effectively represent images across variations of scale and viewpoint changes, clutter background and occlusion~\cite{Unnikrishnan06,miao2013median}. For years, interest point detectors have been extensively studied and widely used in many applications \cite{Tuytelaars08,Guan12,Wu13,Chen13,Miao12icassp}. Nevertheless, an open question remains about extracting the stable points under illumination variations. The Hessian-Laplace/Affine~\cite{Mikolajczyk04}, Harris-Laplace/Affine~\cite{Mikolajczyk04}, SIFT~\cite{Lowe04}  and SURF~\cite{Bay06} detectors are built upon the derivatives of the Gaussian filter. Either the first or the second derivative of the Gaussian filter is used to compute the strength of the image local contrast. As the Gaussian filter responds proportionally to the image local contrast, these detectors perform poorly in detecting low contrast structures even if these structures are stable under different variations and significant in computer vision applications. Moreover, these detectors are susceptible to abrupt structures and image noises. To mitigate the influence caused by image noise and nearby image structures, a rank-ordered Laplacian of Gaussian filter is proposed in \cite{Miao13PR}. However, such a detector still partial relies on the image local contrast.


To address the problems caused by illumination changes particularly, image segmentation has been utilized in designing interest point detectors. For example, the MSER~\cite{Matas04,Kimmel11}, PCBR~\cite{Deng07} and BPLR~\cite{Kim11} detectors use the watershed-like segmentation algorithms to extract the image structures. However, these detectors' performance is unsatisfactory under image blurring in which the boundaries of image structures are unclear~\cite{Tuytelaars08}. Self-dissimilarity and self-similarity of image patches are used in SUSAN \cite{Smith97}, FAST \cite{Rosten10} and self-similar~\cite{Maver10} detectors to alleviate the problems caused by lighting variation. In particular, the SUSAN and FAST detectors use the number of pixels that are dissimilar from that in a region center to detect corners. The weakness of two detectors is that they are not scale-invariant and inefficient in detecting blob-like structures. Although local pixel variance is adopted in~\cite{Maver10} to estimate the self-similarly, the robustness of this detector is uncertain when there are strong abrupt changes within the image patch. 

Considering the above-mentioned limitations of existing detectors, this paper aims to develop a contrast invariant and noise resistant interest point detector. Inspired by the recent work on the Iterative Truncated Mean (ITM) algorithms \cite{Jiang12,Miao13,Miao12,MIAO2014147}, an adaptive ternary coding (ATC) is proposed to adaptively encode the pixels into bright, dark and uncertain statues. The ternary status of each pixel in a local region is detected by the dynamic thresholds that are automatically computed by the ITM algorithm. Interest points are extracted from the blob significance map that is measured by the number of bright and dark pixels. As expected, the proposed ATC shows robustness to illumination variations and is effective in dealing with cluttered structures.

\section{The Proposed Interest Point Detector}


\begin{figure}[!t]
\centering
\subfigure[]{ \includegraphics[width=0.22\textwidth]{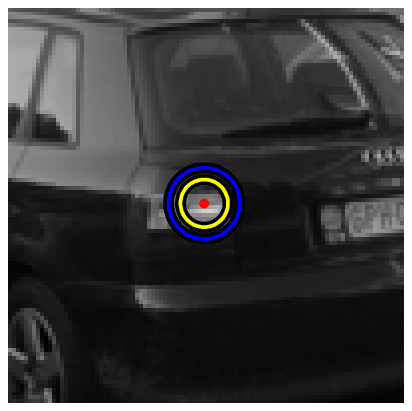} \label{image_patch} }
\subfigure[]{ \includegraphics[width=0.1\textwidth]{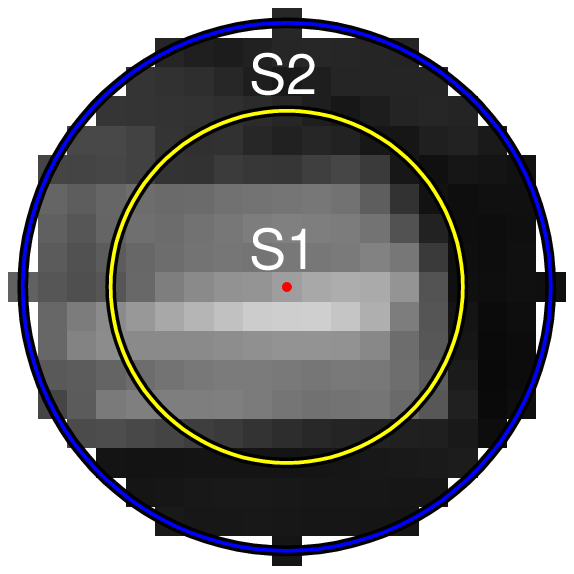} \label{image_patch_zoom} }
\subfigure[]{ \includegraphics[width=0.26\textwidth]{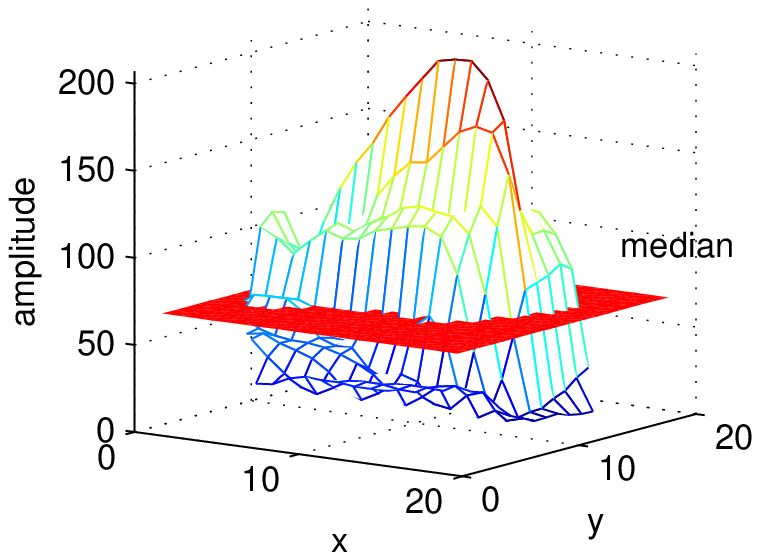} \label{data_median} }
\subfigure[]{ \includegraphics[width=0.26\textwidth]{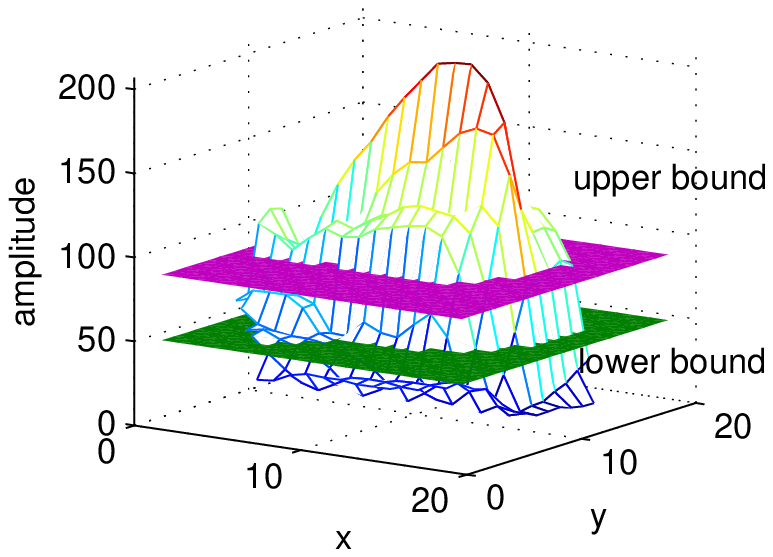} \label{data_upper_lower_bound} }
\subfigure[]{ \includegraphics[width=0.26\textwidth]{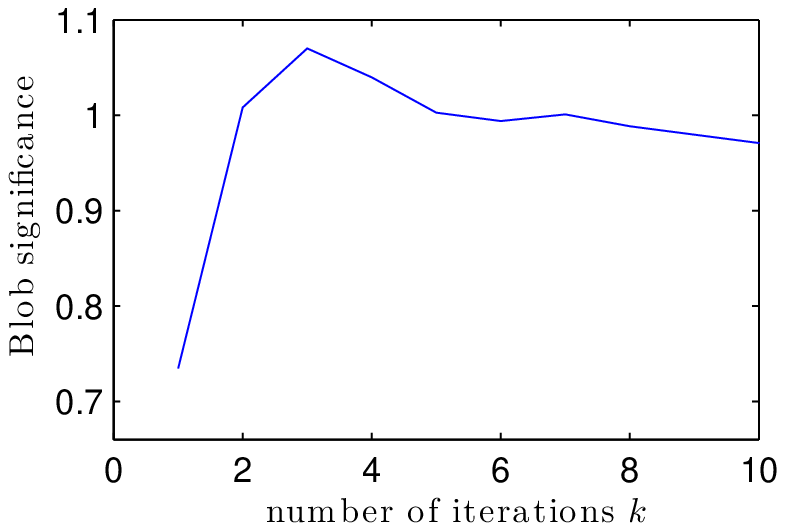} \label{blob_significance} }

\caption{(a) input image. (b) an enlarged image patch. For this image patch, (c) shows the pixels divided by the median and (d) shows the pixels divided by the upper and lower bounds of the ITM algorithm. (f) shows the corresponding blob significance against the number of iterations. (Best viewed in color).}
\label{image_example}
\vspace{-0.3cm}
\end{figure}

\subsection{Problem Formulation}


Blobs, as shown in Fig. \ref{image_patch_zoom},  are the image local structures with the majority of the bright (or dark) pixels concentrating in the center while the majority of opposite intensity  resides in the peripheral region. Such property of the blob structure is preservable under various variations. Moreover, the blob-like structures widely spread over a pictorial image. These properties make the blob-like structure suitable in anchoring the local descriptor~\cite{miao2017laplace,Lowe04} under various image conditions. Hence, a lot of works have been proposed to extract blob-like structures from images \cite{Lowe04,Matas04,Maver10,Miao2015}. However, the linear filter based detectors, such as SIFT and SURF, are sensitive to the illumination changes. In contrast, the relative bright-dark order of pixels in a local region is more stable than the pixel intensity value under illumination changes. In view of this, we propose to detect interest points using the bright/dark labels of pixels.

An issue that needs to be addressed is how to differentiate and label the pixels as bright or dark ones. One way is to dichotomize the pixels into bright and dark ones by a certain threshold, which could be set by the mean or median value of the local region. Take the image patch (shown in Fig. \ref{image_patch_zoom}, as a zoom in from Fig. \ref{image_patch}) as an example, the bright and dark pixels dichotomized by the median value are identified in Fig. \ref{data_median}. Median is more robust to the outliers and abrupt variations than mean. However, the median-based threshold is sensitive  to quantization error because of its inefficiency in suppressing this type of noise. This may lead to unreliable labelling. To solve this problem, we propose to introduce a fuzzy label for the pixels that are not clear enough to be labelled into either bright or dark set. This results in our proposed adaptive ternary coding algorithm.

\subsection{Adaptive Ternary Coding Algorithm}


Instead of using one threshold to binarize the pixels into bright or dark labels, a pixel intensity margin spanned by two thresholds is proposed to ternarize the pixels, as 
\begin{equation}
\label{ternary_label}
\varTheta(I,\lambda_l,\lambda_h) \triangleq\left\{\begin{array}{lll}
 1,& \textrm{if $I\geq \lambda_h$} & (bright)\\
 0,& \textrm{if $\lambda_l<I<\lambda_h$} & (uncertain)\\
 -1,& \textrm{if $I\leq \lambda_l$} & (dark)\\
\end{array}.
\right.
\end{equation}
where $I$ is the pixel intensity value, $\lambda_l$ and $\lambda_h$ are the lower and upper bounds for the pixel ternarization. Pixel intensities that are close to the median value in a local region are labeled the uncertain ones to reduce their sensitivity to noise. Properly choosing the two thresholds is essential in the ternarization. The two thresholds should be invariant to the illumination changes, and should be located on both sides of the median value to ensure the correctness of pixel labeling.

Let the half width of the margin spanned by $\lambda_l$ and $\lambda_h$ be $\tau_\lambda = (\lambda_h-\lambda_l)/2$, and the mean of $\lambda_l$ and $\lambda_h$ be $\mu_\lambda=(\lambda_h+\lambda_l)/2$. Choosing $\lambda_l$ and $\lambda_h$ is equivalent to choosing $\mu_\lambda$ and $\tau_\lambda$. One solution for the ternary coding is setting $\mu_\lambda$ equal to the median of the local region and $\tau_\lambda$ equal to some fixed threshold. However, this has two limitations: 1) computing the median is time consuming and 2) a fixed threshold cannot adapt to the contrast changes. Compared to the median, the mean $\mu$ of the pixel intensities in a local region is easier to be computed. By setting $\tau_\lambda$ equal to the Mean Absolute Deviation (MAD)  $\tau$ of the pixel intensities from the mean $\mu$, the two thresholds $\lambda_l=\mu-\tau$ and $\lambda_h=\mu+\tau$ are located on both sides of the median \cite{Jiang12} and invariant to the illumination changes. Moreover, by iteratively truncating the extreme samples with the ITM algorithm proposed in \cite{Jiang12,Miao13}, the mean of the truncated data starts from the mean and approaches to the median of the input data. Meanwhile, the MAD of the truncated data converges to zero \cite{Jiang12,Miao13}. As a result, these two boundaries $\lambda_l$ and $\lambda_h$ computed by the ITM algorithm automatically converge to the median while keeping the median within the margin spanned by $\lambda_l$ and $\lambda_h$. Therefore, this margin (as shown in Fig. \ref{data_upper_lower_bound}) separates the pixels into bright and dark ones and tolerates noise and quantization errors. Given the advantage of the ITM filter, we propose an adaptive ternary coding algorithm and a blob significance measure based on the ITM algorithm, which are presented as follows.

Let $S_1$ and $S_2$ be the central region and the corresponding peripheral ring of a filter mask centered at $(0,0)$. For the blob detection, here both $S_1$ and $S_2$ are chosen as circle shape, and the radius of the outside ring is $\sqrt{2}$ times of the inner one to make the area size of these two regions the same. Two pixel sets centered at $\mathbf{x}$ are defined as $\mathbf{I}_1(\mathbf{x}) = \{ I(\mathbf{x}-\mathbf{m})| \mathbf{m} \in S_1\}$ and $\mathbf{I}_2(\mathbf{x})=\{I(\mathbf{x}-\mathbf{m})|\mathbf{m}\in S_2\}$, where $\mathbf{x}$ is the region center and $I(\mathbf{x}-\mathbf{m})$ is the pixel gray value at the location $\mathbf{x}-\mathbf{m}$. In order to ensure that the two pixel sets $\mathbf{I}_1(\mathbf{x})$ and $\mathbf{I}_2(\mathbf{x})$ have the same effect on estimating the thresholds for pixel labeling, the weighted ITM algorithm \cite{Miao13} is adopted to make them have equivalently equal number of pixels. The pixel numbers $n_1$ and $n_2$ in these two sets $\mathbf{I}_1(\mathbf{x})$ and $\mathbf{I}_2(\mathbf{x})$ are used to weight the pixels in $\mathbf{I}_2(\mathbf{x})$ and $\mathbf{I}_1(\mathbf{x})$, respectively. The proposed adaptive pixel ternary coding is shown in Algorithm \ref{alg:orig_ITM2}.

\begin{algorithm}[h]
\SetAlgoLined
\LinesNumbered
\KwIn{$\mathbf{I}_1(\mathbf{x})$, $\mathbf{I}_2(\mathbf{x})$, $n=n_1+n_2$, $k=0$; \\
 \textbf{Output}: Blob significance $B(\mathbf{x},k)$;}

\doWhile{the stopping criterion $C_d$ is violated}{
Compute the weighted mean $\mu_w=(n_2\sum\mathbf{I}_1(\mathbf{x})+n_1\sum\mathbf{I}_2(\mathbf{x}))/n$;

Compute the weighted dynamic threshold $\tau_w=(n_2\sum|\mathbf{I}_1(\mathbf{x})-\mu_w|+n_1\sum|\mathbf{I}_2(\mathbf{x})-\mu_w|)/n$;

$k = k+1$, $\lambda_l(k)=\mu_w-\tau_w$, $\lambda_u(k)=\mu_w+\tau_w$, compute the blob significance $B(\mathbf{x},k)$ by (\ref{brightness}),  and truncate $I_i\in\{\mathbf{I}_1(\mathbf{x}),\mathbf{I}_2(\mathbf{x})\}$ by:
\begin{equation}
    I_i=\left\{\begin{array}{ll}
    \lambda_u(k),& \textrm{if $I_i> \lambda_u(k)$}\\
    \lambda_l(k),& \textrm{if $I_i< \lambda_l(k)$}\\
    I_i, & \textrm{otherwise}
     \end{array}\nonumber
     ;\right.
\end{equation}\label{alg:step:ITM2}
}
\caption{Adaptive Pixel Ternary Coding for the Proposed Detector}
\label{alg:orig_ITM2}
\end{algorithm}


The lower and upper bounds $\lambda_l$ and $\lambda_u$ in Algorithm  \ref{alg:orig_ITM2} are used to ternarize the pixels into bright, uncertain or dark ones by (\ref{ternary_label}), as shown in Fig. \ref{data_upper_lower_bound}. A bright pixel is the one that is larger than the higher threshold. A dark pixel is the one that is smaller than the lower threshold. The blob structures have the attribute that the majority of bright (or dark) pixels are concentrated in the inner region while the majority of the opposite ones in the surrounding region.  As a result, we measure the blob significance by the distribution of the bright and dark pixels. First, the dominances of bright/dark pixels in $S_1$ and $S_2$ are measured by the difference of the numbers of bright and dark pixels in the corresponding region. The bright and dark pixels are respectively labeled as $1$ and $-1$ by (\ref{ternary_label}) and the uncertain pixels are labeled as 0. Therefore, the normalized dominance of the bright/dark pixels in $S_1$ and $S_2$ are $\frac{1}{n_1}\sum\varTheta(\mathbf{I}_1(\mathbf{x}),\lambda_l(k),\lambda_h(k))$ and $\frac{1}{n_2}\sum\varTheta(\mathbf{I}_2(\mathbf{x}),\lambda_l(k),\lambda_h(k))$, respectively, where $\lambda_l(k)$ and $\lambda_h(k)$ are the lower and upper bounds in the $k$th iteration. Second, these two parts are linearly combined as the blob significance in the $k$th iteration:
\begin{align}
\label{brightness}
B(\mathbf{x},k) &= \frac{1}{n_1}\sum\varTheta(\mathbf{I}_1(\mathbf{x}),\lambda_l(k),\lambda_h(k))\nonumber\\
&-\frac{1}{n_2}\sum\varTheta(\mathbf{I}_2(\mathbf{x}),\lambda_l(k),\lambda_h(k)).
\end{align}
From Algorithm \ref{alg:orig_ITM2} it is seen that the margin between the lower and upper bounds equals $2\tau_w$. It monotonically decreases to zero by increasing the number of iterations \cite{Miao13}. In the first few iterations, the margin is large as only few extreme samples are truncated by the ITM algorithm. By increasing the number of iterations, both the lower and higher thresholds converge to the median value of the local region. As a result, the margin between these two thresholds reduces. Therefore, the number of pixels categorized into the intermediate group decreases. The blob significance $B(\mathbf{x},k)$ (shown in Fig. \ref{blob_significance}) is a function of the number of iterations $k$. The maximum value of $|B(\mathbf{x},k)|$ over $k$ is selected as the blob significance map for interest point detection, defined as
\begin{equation}
\label{blob_significant_region}
B(\mathbf{x})  \triangleq B(\mathbf{x},k) \textrm{ with } k = \argmax_k(|B(\mathbf{x},k)|).
\end{equation}
However, exhaustively searching the global peaks over all iterations is time-consuming. The following stopping criterions are used to allow that the global maximum value is achieved in most cases within a reasonable number of iterations.

Let $\mathbf{I}(\mathbf{x})=\mathbf{I}_1(\mathbf{x})\cup\mathbf{I}_2(\mathbf{x})$, the corresponding weight set be $\mathbf{w}=\{\underbrace{n_2,...,n_2}_{n_1 \textrm{ times}},\underbrace{n_1,...,n_1}_{n_2 \textrm{ times}}\}$ and the two sets separated by the weighted mean $\mu_w$ be $\mathbf{I}_h(\mathbf{x})\triangleq\{I_i|I_i\in\mathbf{I}(\mathbf{x}),I_i>\mu_w\}$ and $\mathbf{I}_l(\mathbf{x})\triangleq\{I_i|I_i\in\mathbf{I}(\mathbf{x}),I_i\leq\mu_w\}$. Let $w_h$ and $w_l$ denote the summation of the weights of $\mathbf{I}_h(\mathbf{x})$ and $\mathbf{I}_l(\mathbf{x})$, respectively. One stopping criterion \cite{Miao13}, which enables the truncated mean to be close to the weighted median, is to meet the condition
\begin{equation}
C_1: |w_h-w_l|\leq \max\{n_1, n_2\}.
\end{equation}
In some cases, after $C_1$ is met, the amplitude of the blob significance $B(\mathbf{x},k)$ still increases  because the number of pixels with  uncertain status is still large. Therefore, an additional constrain is applied:
\begin{equation}
C_2: |B(\mathbf{x},k)|\leq |B(\mathbf{x},k-1)|.
\end{equation}
The third condition is to limit the maximum number of iterations as
\begin{equation}
C_3: k\geq 2\sqrt{n},
\end{equation}
which is chosen from experiment. 
The truncating procedure of in Algorithm \ref{alg:orig_ITM2} is terminated if the following conditions is satisfied, as
\begin{equation}
C_d: (C_1\wedge C_2) \vee C_3.
\end{equation}

\begin{figure*}[!t]
\centering
\subfigure[]{ \includegraphics[width=0.181\textwidth]{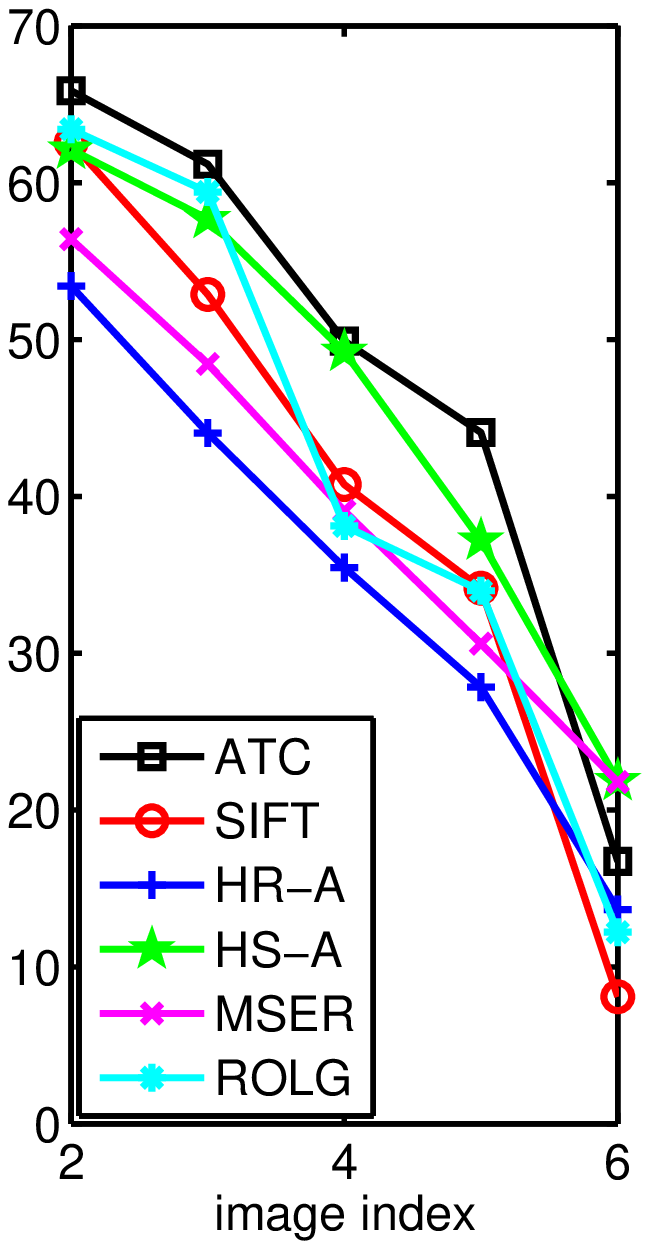} \label{repeatability_wall} }
\subfigure[]{ \includegraphics[width=0.181\textwidth]{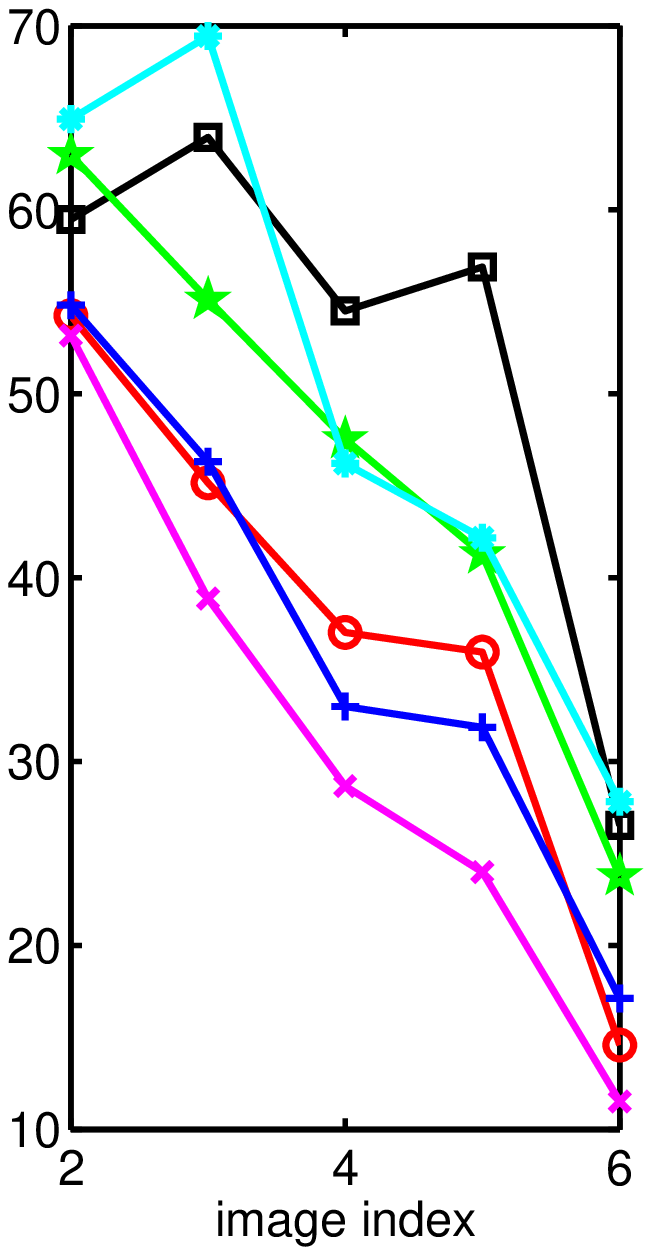} \label{repeatability_boat} }
\subfigure[]{ \includegraphics[width=0.181\textwidth]{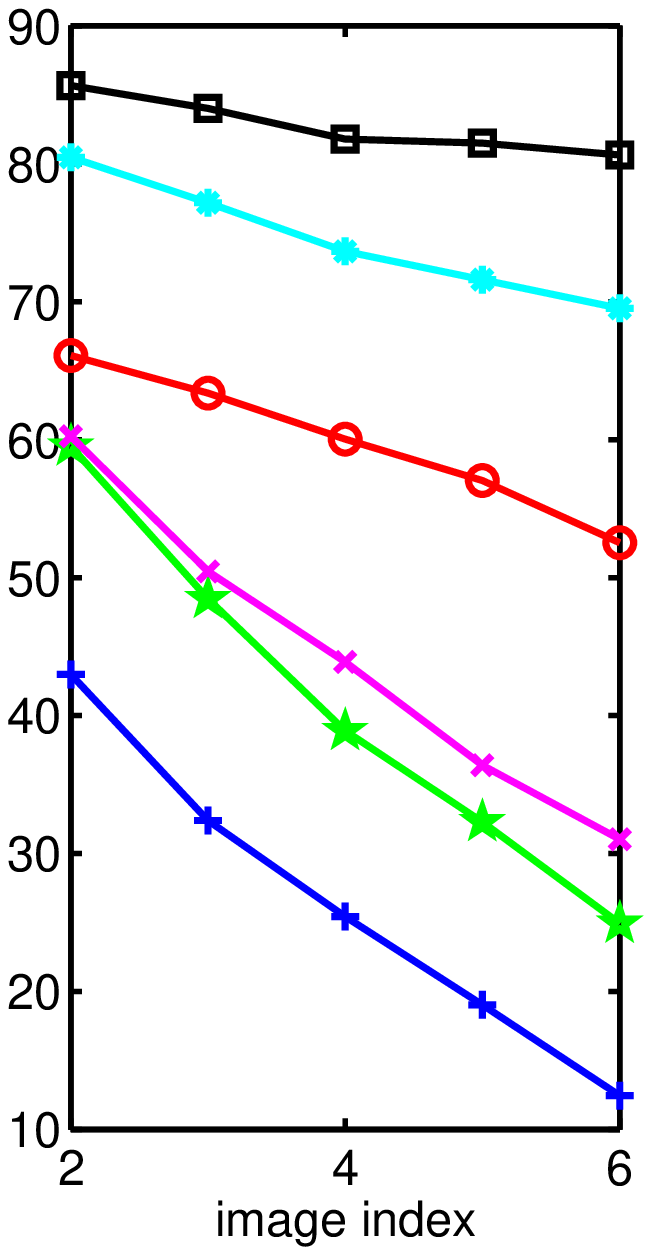} \label{repeatability_tree} }
\subfigure[]{ \includegraphics[width=0.181\textwidth]{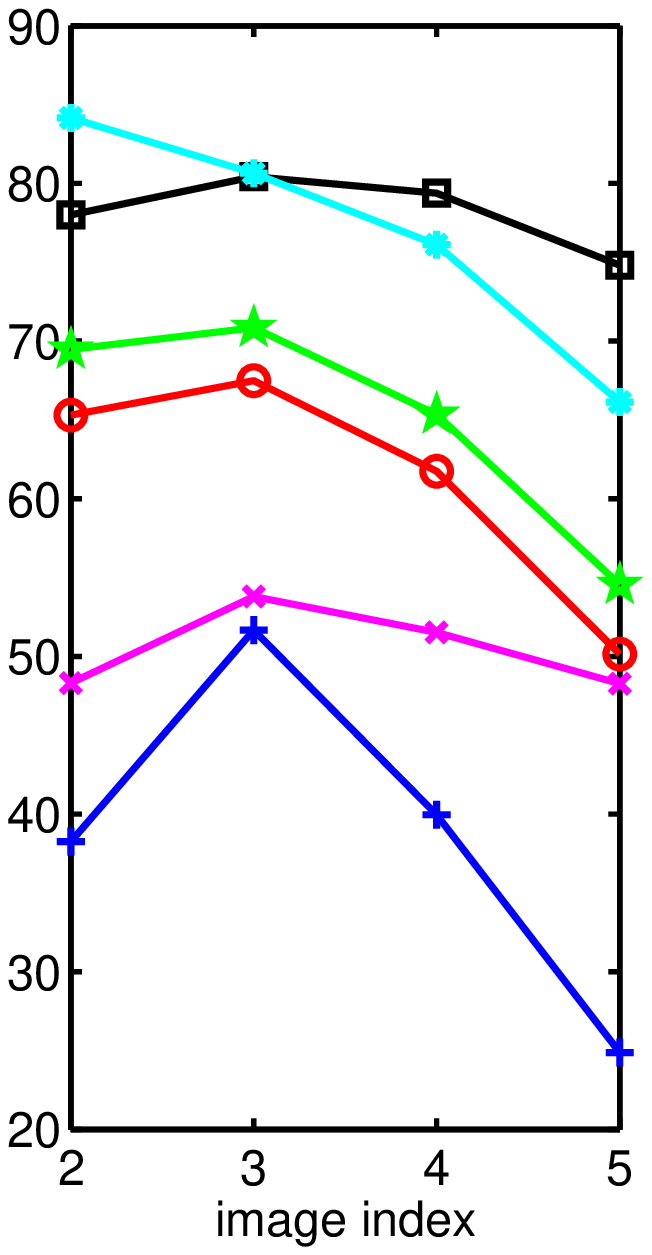} \label{repeatability_desktop} }
\subfigure[]{ \includegraphics[width=0.181\textwidth]{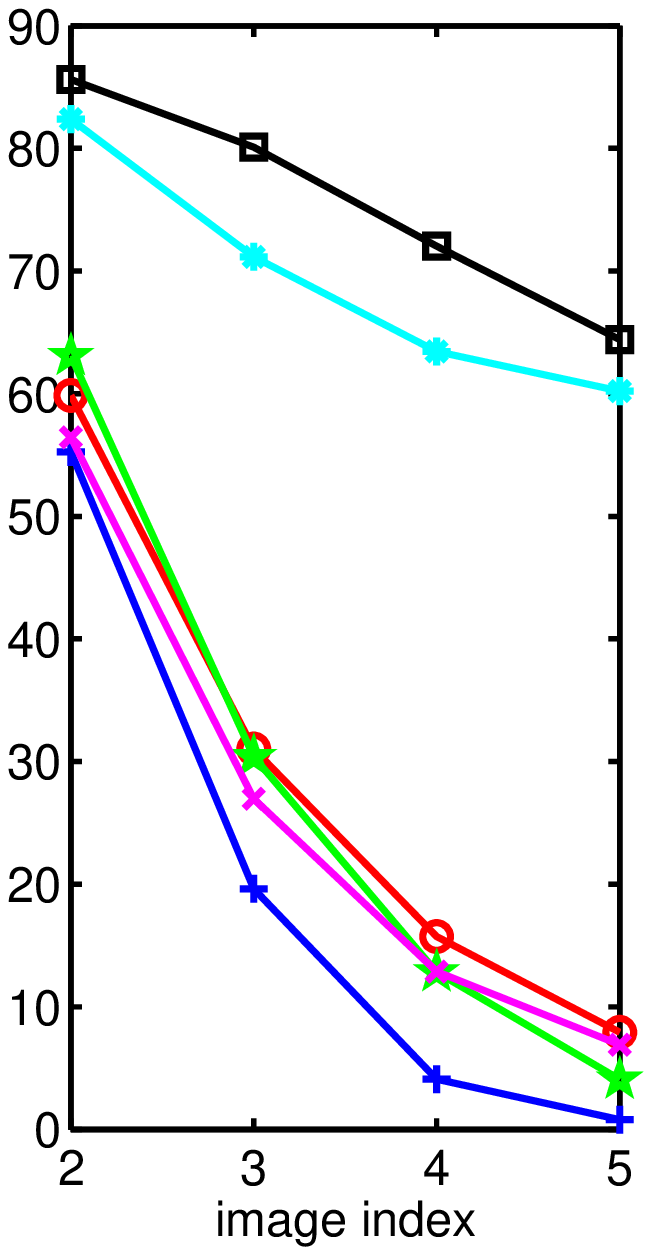} \label{repeatability_corridor} }

\caption{Results on (a)  textured scene `wall' v.s. viewpoint angle changes from 20 degree to 60 degree, (b) `boat' structured sequence  v.s.  scale changes from 1.1 to 2.8, (c) `leuven' the illumination change sequence with decreasing light, (d)`desktop' and (e) `corridor' with complex illumination changes.}
\label{repeatability}
\vspace{-0cm}
\end{figure*}

From (\ref{brightness}) we find that the blob significance value $B(\mathbf{x})$ is within the range $[-2, 2]$. For a bright region, $B(\mathbf{x})>0$. The maximum value of its blob significance is 2. Similarly, a local region is dark if $B(\mathbf{x})<0$ and the minimum value of its blob significance is -2.

\subsection{The Proposed ATC Detector}

\subsubsection{Ridge and Edge Suppression}

Interest points are extracted by detecting the local peaks from the blob significance map (\ref{blob_significant_region}). In order to suppress the unreliable points detected on ridges and edges,  the ratio
\begin{equation}
\label{keypoint_remove}
r = \frac{|B(\mathbf{x})|-\max\{|B(\mathbf{x}-\mathbf{m})||\mathbf{m}\in S_2\}}{\max\{|B(\mathbf{x}-\mathbf{m})||\mathbf{m}\in S_2\}}
\end{equation}
is used. Small $r$ means that the peak value is quite similar to that in its surrounding regions. We remove such candidates if $r<0.05$, which is chosen empirically.

\subsubsection{Algorithm for ATC Detector}

Detecting interest points in multiple scales is essential in many vision applications where the same objects can appear with different sizes. By changing the size of the local image patches $S_1$ and $S_2$, the ATC detector can identify local structures of various scales. Similar to that done in \cite{Lee09}, we implement the multi-scale ATC detector by detecting the points in each scale. The procedures of the proposed ATC detector are summarized as follows:
\begin{enumerate}
\item Generate the blob significance map on multi-scales by Algorithm \ref{alg:orig_ITM2}. 
\item Detect the local peaks of the blob significance on spatial dimensions.\label{code2}
\item Remove the peaks on ridges and edges by (\ref{keypoint_remove}). The remaining peaks are the interest points to be detected.
\end{enumerate}

\begin{table}[t]
\caption{Number of Detected Points on the First Image of Each Data Set.}
\label{tab_oxford_points}
\begin{center}
\begin{tabular}{@{\hspace{\tabcolsep}%
                \extracolsep{\fill}}ccccccc} \hline
           & ATC  			& SIFT     &   HR-A    & HS-A     	& MSER  & ROLG\\\hline
wall       & 1508   		& 1460   	&  1520	    & 1568	 	& 1593  & 1514\\\hline
boat       & 1546     		& 1501     	&  1549   	& 1429 		& 1524  & 1501\\\hline
leuven     & 1527   		& 1426   	&  1476     & 1501  	& 1648  & 1488\\\hline  
desktop    & 1539   		& 1539   	&  868      & 1526   	& 1698  & 1451 \\\hline
corridor   & 1526     		& 1564 		&  1540		& 1544  	& 1583  & 1578 \\\hline

\end{tabular}
\vspace{-0.0cm}
\end{center}
\end{table}

\section{Experiments}

\subsection{Repeatability}

Two detected regions are regarded as repeated if their overlap is above 60\% as suggested in~\cite{Mikolajczyk05T}. For an image pair \{\textrm{Img1}, \textrm{Img2}\}, the repeatability score  is defined as  $p_r/\max\{p_1,p_2\}$, where $p_r$ is the number of repeated points, and $p_1$ and $p_2$ are the numbers of the points detected from the common area and scale of $\textrm{Img1}$ and $\textrm{Img2}$, respectively. We use the repeatability to evaluate the detectors under different variations.  The three datasets `wall', `boat' and `leuven' from Oxford database in \cite{Mikolajczyk05T} and the `desktop' and `corridor' datasets from \cite{Wang11} with complex illumination changes are used for testing.

Similar to that in \cite{Maver10}, half-sampled images are used for  evaluation. For the ATC detector, interest points are extracted on 5 octaves by half-sampling the previous octave. In each octave, local extrema are detected on 3 scales: $\{\sigma_n\}_{n=1, 2, 3}=\{ 4, 5, 6\}$. The ATC detector is compared with five detectors consisting of  the SIFT~\cite{Lowe04}, Harris-affine (HR-A)~\cite{Mikolajczyk04}, Hessian-affine (HS-A)~\cite{Mikolajczyk04}, MSER~\cite{Matas04} and ROLG \cite{Miao13PR} detectors.  For each data set, the detector parameters are adjusted so that roughly the same number of interest points (shown in Table \ref{tab_oxford_points}) are detected on the first image for all detectors. The interest points detected by the HR-A detector on the first image of the `desktop' set is smaller than others although the contrast threshold is already set to be zero due to the darken illumination on this image. Fig. \ref{repeatability} (a) and (b) illustrate the experimental results under the changes of viewpoint and scale, respectively. Fig. \ref{repeatability} (c), (d) and (e) show the performances under complex illumination changes. These results show that the ATC detector can achieve better performance than the other five detectors under almost all the different experimental settings.

\subsection{Application to Face Recognition}

To demonstrate the implications of the proposed ATC detector, we evaluate it in the face recognition application \cite{Jiang08,Miao08,miao2009human}. Specifically, the ATC detector is compared with the SIFT~\cite{Lowe04}, HR-A~\cite{Mikolajczyk04}, HS-A~\cite{Mikolajczyk04}, MSER~\cite{Matas04} and ROLG \cite{Miao13PR} detectors. As the default setting produces too few interest points for the face recognition for all detectors, the thresholds that are used to remove the low response interest points are set to be zero for all detectors in the present experiment. For the MSER detector, the minimum size of its output region is set to be 1/4 of the default setting to ensure it is applicable to all of the testing databases. All the detected interest points are described by the SIFT descriptor. The matching algorithm for face recognition, which consists of interest point matching and geometric verification with Hough transform, is described in~\cite{Lowe04}.


Four standard face recognition databases, including AR~\cite{Martinez02},  GT~\cite{Georgia07},  ORL~\cite{Samaria94} and FERET~\cite{Phillips00}, are used to evaluate these detectors. The database setting is shown in Table \ref{tab_database_setting}. The face images in these databases have variations in illumination, expression and poses. The recognition rate, which is the percentage of correctly identified test images from the rank-1 best matched gallery, is used to measure the performance of the interest point detectors. Table \ref{tab1} shows that the proposed detector achieves the highest recognition rate over the four databases. It suggests that the interest points detected by the proposed ATC detector are more robust and discriminative compared to others.

\begin{table}[t]
\caption{Face Database Settings.}
\vspace{-0.0cm}
\label{tab_database_setting}
\begin{center}
\begin{tabular}{@{\hspace{\tabcolsep}%
                \extracolsep{\fill}}cccccc} \hline
                 & image size          & subjects          &   gallery         &  test  \\\hline
AR               & 60$\times$85        &75        &  7     &  7 \\\hline
GT               & 60$\times$80       &50        & 8   &  7  \\\hline
ORL              & 50$\times$57      &40           & 5     & 5 \\\hline
FERET            & 60$\times$80     &1194       & 1   &  1 \\\hline
\end{tabular}
\end{center}
\vspace{-0.0cm}
\end{table}

\begin{table}[t]
\caption{Recognition Rate on AR, GT, ORL and FERET Databases.}
\vspace{-0.0cm}
\label{tab1}
\begin{center}
\begin{tabular}{@{\hspace{\tabcolsep}%
                \extracolsep{\fill}}cccccc} \hline
                         & AR    &    GT         & ORL                  & FERET  \\\hline
ATC             & \textbf{98.3\%}    &  \textbf{94.0\%}    &\textbf{97.5\%}              &  \textbf{98.5\%} \\\hline
 SIFT           & 94.3\%      & 84.0\%     &90.0\%                &89.9\% \\\hline
 HS-A           & 88.6\%      & 74.0\%     &80.0\%           &  85.3\%  \\\hline
 HR-A           & 74.5\%      & 47.4\%   &66.5\%         &  49.7\% \\\hline
 MSER          & 92.7\%       & 81.1\% & 91.0\%          &  89.3\% \\\hline
ROLG            & 98.3\%      & 91.1\%     &96.5\%                &98.2\% \\\hline
\end{tabular}
\end{center}
\vspace{-0.0cm}
\end{table}


\section{Conclusions}

In this paper, an interest point detector is designed based on the adaptive ternary coding (ATC) algorithm, which is inspired by the ITM algorithm to categorize the pixels into the bright, dark and uncertain statuses. As the blob significance is measured by counting the number of bright and dark pixels, the detection result is invariant to the illumination changes.  Evaluations on the Oxford dataset \cite{Mikolajczyk05T} and the complex illumination dataset in \cite{Wang11} show that the ATC detector outperforms the other five detectors in terms of repeatability under the variations caused by scale, viewpoint and illumination changes. The advance performance of the proposed detector is also verified in the application of face recognition.


\bibliographystyle{IEEEbib}
\bibliography{ICME_MyReference}

\end{document}